\pdfoutput=1
\documentclass{article}
\usepackage{iclr2027_conference,times}
\iclrfinalcopy
\usepackage{amsmath,amsfonts,bm}

\def\eqref#1{equation~\ref{#1}}
\def\1{\bm{1}}

\DeclareMathAlphabet{\mathsfit}{\encodingdefault}{\sfdefault}{m}{sl}
\SetMathAlphabet{\mathsfit}{bold}{\encodingdefault}{\sfdefault}{bx}{n}

\usepackage{hyperref}
\usepackage{url}
\usepackage{booktabs}
\usepackage{multirow}
\usepackage{amsmath,amssymb}
\usepackage{graphicx}
\usepackage{float}
\usepackage{xcolor}
\usepackage{enumitem}
\hypersetup{hidelinks}
\newcommand{\ea}{E_{\mathrm{A}}}
\newcommand{\ev}{E_{\mathrm{V}}}
\newcommand{\si}{\mathrm{SI}}

\makeatletter
\newcommand{\tablecaption}{\def\@captype{table}\caption}
\makeatother

\title{Seeing What Should Be Heard:\\Diagnosing and Repairing\\Cross-Modal Shortcuts in Omni-Modal LLMs}

\author{Yueran Ma \\
The University of Queensland, Brisbane, Australia \\
\texttt{s4931778@student.uq.edu.au}
\And
Ronghao Lin \\
Shenzhen University, Shenzhen, China}

\begin{document}

\maketitle
\lhead{Preprint}
\suppressfloats[t]

\begin{abstract}
Omni-modal large language models (LLMs) are expected to answer a question using the modality it explicitly refers to.
However, existing training paradigms rarely verify whether models actually follow this modality, because multimodal inputs from the same sample often provide redundant evidence for the same answer.
In this work, we uncover a pervasive \emph{cross-modal shortcut} in omni-modal LLMs: when asked an audio-related question, models rely on the image as much as on the audio, and sometimes even more.
To systematically diagnose this behavior, we introduce the \emph{Factorized Modality Diagnostic}, which independently swaps audio and images between samples to isolate each modality's causal contribution.
Across two model families in different settings, we find that this shortcut persists throughout supervised fine-tuning and reinforcement learning post-training, while judge-based RL may further amplify such reliance on irrelevant visual information.
Based on this finding, we propose \emph{DMC-Repair}, which trains models on the same kind of cross-modal swapped samples while assigning supervision according to the modality specified by the question.
This prevents models from exploiting the spurious correspondence between modalities within the same clip.
Experiments demonstrate that DMC-Repair reduces the image-induced share of the answer effect by $59.9\%$, effectively suppressing the cross-modal shortcut without compromising audio-question answering performance.
The reduction in shortcut reliance generalizes across two model families and zero-shot to an unseen dataset and an unseen benchmark, and persists through subsequent post-training.
Code is available at \url{https://anonymous.4open.science/r/DMC-Repair}.
\end{abstract}

\section{Introduction}
\label{sec:intro}

\begin{figure}[t]
\centering
\includegraphics[width=\textwidth]{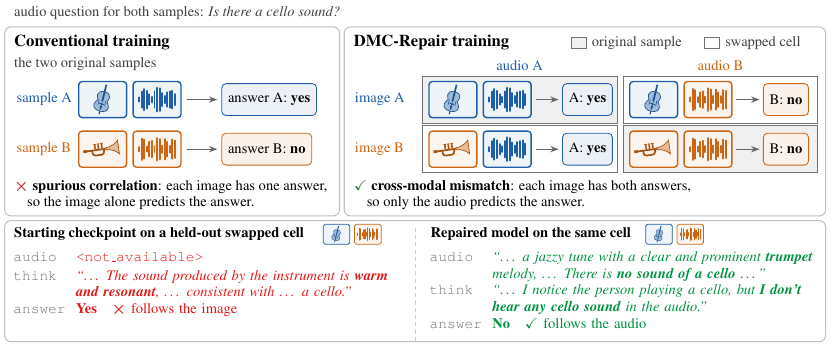}
\caption{Conventional training and DMC-Repair training for an audio question.
DMC-Repair trains on all four image--audio combinations of two samples and labels each cell by its audio.
The bottom rows quote excerpts of the reasoning, with errors in red and audio evidence in green.}
\label{fig:teaser}
\end{figure}

When an omni-modal large language model (LLM) answers a question about the audio of a video, its answer should come from what it hears.
Recent work post-trains such models with reinforcement learning (RL) \citep{humanomniv2,r1omni}, which has improved reasoning in language models \citep{deepseekr1}.
These pipelines reward a correct answer and a well-formed response, and some also use a judge model to score the reasoning text.
However, none of these rewards checks which modality the model actually uses to produce its answer.
Because the image and the audio of a training sample usually support the same answer, a model that is asked which instrument is playing can identify the instrument from the image and still receive the full reward.
We call this behavior a \emph{cross-modal shortcut}.
In the example of Figure~\ref{fig:teaser}, a model that sees a cello while a trumpet plays describes a cello sound and answers from the image.

As in this example, the shortcut often manifests as cross-modal hallucination, in which a model reports a sound only because its source is visible \citep{avhbench}.
Existing methods address this problem in three ways.
Decoding methods contrast or reweight modality-specific branches at inference time \citep{avcd,fmd,mad}.
Preference optimization favors grounded responses over hallucinated ones \citep{acpo,omnidpo,moddpo}, and recent RL methods make the training objective modality-aware \citep{sffl,omnivideo_r1,mapo}.
However, these methods require extra decoding passes, preference pairs, or additional rewards.
They are also evaluated by benchmark accuracy, and they rarely verify which modality an answer actually relies on.

Such verification needs cases in which the image and the audio disagree, but natural data rarely contains them.
We therefore create them by swapping the audio and the image between real samples and call the resulting test the Factorized Modality Diagnostic.
For an audio question, the diagnostic takes two samples with different answers and forms a grid of the four combinations of their images and audio, which we call \emph{cells}.
In each cell, the \emph{answer margin} is the gap between the log-probabilities of the two answers, and comparing the four margins separates the audio effect from the image effect.
The \emph{Shortcut Index} is the image's share of the answer effect and should be zero on audio questions.
On existing omni-modal LLMs, however, the diagnostic reveals this shortcut.
In two model families, the Shortcut Index is close to one half, so the image changes the answer to an audio question about as much as the audio does.
The index remains at this level throughout supervised fine-tuning (SFT) and RL post-training, and judge-based RL rewards designed to encourage modality use further increase the image effect.
As a result, on \emph{conflict cells}, where the two modalities support different answers, the answer to an audio question often follows the image.

The same grid that reveals the shortcut also provides the training signal to suppress it.
When a question designates the audio, we label each cell with the answer that its audio supports, so each answer appears in both original and swapped cells (Figure~\ref{fig:teaser}).
A model therefore cannot infer the answer from the image or from whether the image and the audio originate from the same clip.
We train on these \emph{designated-modality counterfactuals} (DMC) and supervise only the answer tokens.
We call this method DMC-Repair, and it requires no judge, reward model, or preference pairs.

We evaluate the repair with two metrics.
The Shortcut Index shows how much the image drives the answer to an audio question, and \emph{audio-following}, the rate at which generated answers on conflict cells follow the audio, shows whether the model relies on the audio.
On held-out test questions, DMC-Repair reduces the Shortcut Index by $59.9\%$ and improves audio-following by $13.7$ points.

\textbf{Contributions.}
\begin{itemize}[leftmargin=1.1em,nosep,before=\vspace{-\parskip}]
\item We introduce the Factorized Modality Diagnostic, which isolates the effect of each modality on an answer, and reveal that the cross-modal shortcut persists across models and post-training stages.
\item We propose DMC-Repair, which supervises only the answer tokens of counterfactual grids relabeled by the designated modality and suppresses the shortcut better than simpler constructions.
\item Extensive experiments demonstrate that the repair is also effective on another model family, generalizes zero-shot to an unseen dataset and an unseen benchmark, and persists through subsequent RL post-training.
\end{itemize}

\section{Related Work}
\label{sec:related}

\textbf{Reinforcement learning for omni-modal reasoning.}
RL post-training has been applied across multimodal reasoning, from structured-context pipelines \citep{humanomniv2} to video and emotion reasoning \citep{videor1,r1omni}, generally using GRPO or its variants \citep{grpo,dapo,drgrpo}.
These pipelines add a judged context term \citep{humanomniv2} or a frame-order term \citep{videor1} to the answer and format rewards.
None of these rewards verifies whether an answer follows the designated modality.
\citet{omnir1_audio} report GRPO gains in audio QA that persist when the audio input is removed.
Recent work adds modality-aware terms to the objective \citep{sffl,omnivideo_r1,mapo}.
We also add our own modality rewards to RL, and the diagnostic shows that none of them removes the shortcut (Section~\ref{sec:rewards_body}).
DMC-Repair instead changes which input predicts the training label.

\textbf{Diagnostics for modality dependence.}
Media interventions are a common diagnostic, from swap, mute, and shift interventions that expose failures in audio-visual reasoning \citep{wvss} to modality shuffling that shows visual dominance \citep{sensorypid}.
Our diagnostic complements them by measuring each modality's effect on the answer margin with a pair of real samples.
It tests whether the non-designated modality has zero effect for audio and visual questions alike.

\textbf{Counterfactual media as training signal.}
Beyond diagnosis, \citet{wvss} tune models on their interventions to verify audio-visual consistency and describe mismatches, and preference methods build preference pairs from perturbed media \citep{acpo,moddpo,omnidpo}.
Closest to our setting, \citet{somemodalities} test whether answers follow the requested modality on clips with swapped audio and fine-tune on aligned and misaligned clips to answer the category that the question names.
Our construction completes each sample pair into a $2\times2$ grid for one question, so mismatch status carries no information about the answer.
The same grid yields both the training cells and the diagnostic.
Appendix~\ref{app:related} discusses these lines of work further.

\section{Cross-Modal Shortcuts in Omni-Modal LLMs}
\label{sec:audit}

\begin{figure}[t]
\centering
\includegraphics[width=\textwidth]{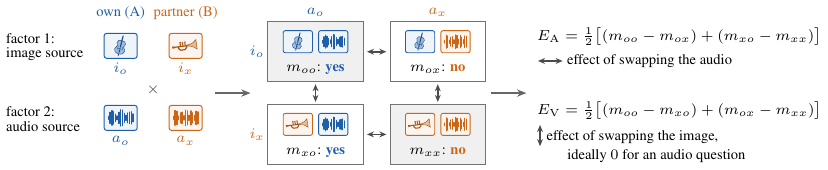}
\caption{The $2\times2$ factorized design for an audio question.
Crossing the image source and the audio source, each with two levels, own (A, blue) and partner (B, orange), gives four cells with answer margins $m_{uv}$ for image $u$ and audio $v$.
Shaded cells contain one sample's image and audio.}
\label{fig:grid}
\end{figure}

\subsection{Problem Setup}
\label{sec:setup}

To formalize the shortcut, we consider an omni-modal model that receives $N$ modality streams $M = (M_1,\dots,M_N)$ together with a question $Q$ and defines a distribution $p(A \mid Q, M)$ over answers.
In many audio-visual QA benchmarks, each question is annotated with the subset of streams that determines its ground-truth answer. We write this \emph{designated} subset as $d(Q)$ and its complement as $\bar{d}(Q)$.
A model uses the designated modality when its answer depends only on $M_{d(Q)}$:
\begin{align}
\label{eq:invariance}
p(A \mid Q, M_{d(Q)}, M_{\bar{d}(Q)}) \;=\; p(A \mid Q, M_{d(Q)}, M'_{\bar{d}(Q)})
\quad\text{for every } M'_{\bar{d}(Q)} .
\end{align}
Here, $N=2$, and $M = (i, a)$ consists of an image and an audio clip.
Accuracy on a natural corpus cannot verify Equation~\ref{eq:invariance}, because $M_{d(Q)}$ and $M_{\bar{d}(Q)}$ are correlated there and a cross-modal shortcut incurs no penalty.
We therefore break this correlation by swapping media between samples.

\subsection{The Factorized Modality Diagnostic}
\label{sec:design}

The diagnostic applies to any audio-visual QA dataset in which several samples share a question.
We call a question \emph{Audio}, \emph{Visual}, or \emph{Audio-Visual} according to whether $d(Q)$ is the audio, the image, or both.
For a question $q$ we select two real samples with different ground-truth answers.
The \emph{own} item has media $(i_o, a_o)$ and answer $g$, and the \emph{partner} item has media $(i_x, a_x)$ and answer $g'$.
The design has two factors, the source of the image and the source of the audio, each with two levels, own or partner (Figure~\ref{fig:grid}). We evaluate all four combinations, which we call \emph{cells}, so the two factors are fully crossed as in a classical $2\times2$ factorial experiment \citep{fisher1935}.
The cells $(i_o, a_o)$ and $(i_x, a_x)$ contain a sample's own image and audio and are \emph{matched}, and the other two are \emph{mismatched}.
For an Audio question the correct answer for every cell is the one that its audio supports, so swapping the image should leave the preferred answer unchanged and swapping the audio should reverse it.
In cell $(i_u, a_v)$, the teacher-forced answer margin $m_{uv} = \log p(g \mid q, i_u, a_v) - \log p(g' \mid q, i_u, a_v)$ compares the two answers under direct answer scoring, using the prompt followed by \texttt{<answer>}. Its sign indicates which answer the model prefers.
The audio and image main effects are
\begin{align}
\ea = \tfrac{1}{2}\bigl[(m_{oo} - m_{ox}) + (m_{xo} - m_{xx})\bigr], \quad
\ev = \tfrac{1}{2}\bigl[(m_{oo} - m_{xo}) + (m_{ox} - m_{xx})\bigr],
\end{align}
with interaction $\psi = \tfrac{1}{2}(m_{oo} - m_{ox} - m_{xo} + m_{xx})$ (Appendix~\ref{app:interaction}).
Each main effect averages the change from swapping one modality over both levels of the other, and $\psi$ measures how much that change depends on the other modality. Swapping or removing one modality at a time would measure its effect at a single level of the other only, and could not separate a main effect from an interaction.
Equation~\ref{eq:invariance} applied to an Audio question requires $\ev = 0$, so we summarize each configuration by the \emph{Shortcut Index}
$\si = |\ev| / (|\ea| + |\ev|)$, which is zero for a model that ignores the image and one half when the two modalities contribute equally.
On Visual questions the two modalities swap roles, and the ideal index is one.
Unlike accuracy, the index exposes a violation of Equation~\ref{eq:invariance}.

A \emph{question family} groups the samples that share one question.
We draw two sample pairs from each family and orient each pair so that the own item has the answer \texttt{yes}.
We compute $\si$ from the main effects averaged over a family's two pairs and report its mean over families.

\subsection{Pilot Study}
\label{sec:si_table}

\begin{figure}[t]
\centering
\begin{minipage}[b]{0.40\textwidth}
\centering
\includegraphics[width=\linewidth]{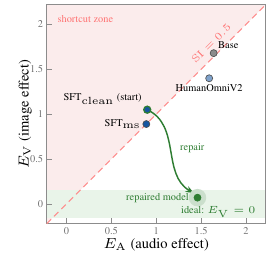}
\end{minipage}\hfill
\begin{minipage}[b]{0.57\textwidth}
\centering
\includegraphics[width=\linewidth]{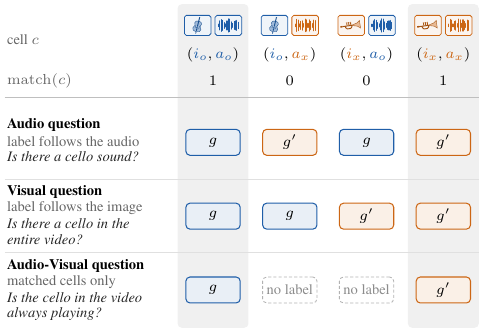}
\end{minipage}
\par\noindent
\begin{minipage}[t]{0.40\textwidth}
\caption{Audio and image effects on Audio questions under the native prompt, which retains the training system prompt and output format (65 development families).}
\label{fig:audit}
\end{minipage}\hfill
\begin{minipage}[t]{0.57\textwidth}
\caption{Labels of the four cells of a pair under Equation~\ref{eq:relabel}, with an example question per type in italics. Blue $g$ is the own sample's answer and orange $g'$ the partner's. Shaded columns are the matched cells ($\mathrm{match}(c) = 1$), which contain one sample's image and audio.}
\label{fig:relabel}
\end{minipage}
\end{figure}

We first run the diagnostic as a pilot study on a development set of MUSIC-AVQA \citep{musicavqa}, with checkpoints from two training routes.
HumanOmniV2 \citep{humanomniv2} adds SFT and GRPO post-training to Qwen2.5-Omni \citep{qwen25omni}, and we diagnose its released checkpoint.
Our repair route starts from SFT$_{\mathrm{clean}}$, an SFT checkpoint trained on HumanOmniV2's cold-start data with separate audio and visual descriptions, and we call it the \emph{starting checkpoint}.
We apply \mbox{DMC-Repair} to SFT$_{\mathrm{clean}}$ to obtain the \emph{repaired model} and continue with GRPO to obtain the \emph{RL endpoint} (Section~\ref{sec:retention}).
We also diagnose SFT$_{\mathrm{ms}}$, which uses an earlier version of these descriptions.

The pilot study finds the same cross-modal shortcut in every checkpoint of both model families, including MiniCPM-o-2.6 \citep{minicpmv,minicpmo}.
Regardless of the prompt and the post-training stage, the image changes the answer about as much as the audio does (Figure~\ref{fig:audit} and Table~\ref{tab:si} in Appendix~\ref{app:audit_full}).
In free generation, the starting checkpoint follows the audio on fewer than half of the conflict cells.
Validity checks confirm that the diagnostic measures modality use rather than noise (Appendix~\ref{app:audit_full}).
The tested SFT and RL post-training stages do not teach a model which modality a question refers to.
On natural samples, an accuracy reward cannot resolve this, because an image-derived answer receives the same reward as an audio-derived one.

\section{DMC-Repair}
\label{sec:method}

The pilot study shows that the model needs training examples in which the image and the audio disagree and the label follows the designated modality.
Our key idea is to build these examples from the same factorized grid that the diagnostic uses.
DMC-Repair consists of three components.
Counterfactual grid construction creates the conflict between the modalities (Section~\ref{sec:cells}), designated-modality relabeling makes the designated modality the only predictor of the label (Section~\ref{sec:relabel}), and answer-token supervision optimizes the answer margin that the diagnostic measures (Section~\ref{sec:supervision}).
The recipe requires only the question types and answers of a dataset (Section~\ref{sec:general}).

\subsection{Counterfactual Grid Construction}
\label{sec:cells}

Let $\mathcal{D}$ be a training corpus of real samples $(q_n, i_n, a_n, y_n)$, where $q_n$ is the question, $i_n$ the image, $a_n$ the audio clip, and $y_n$ the ground-truth answer.
Training first requires inputs in which the two modalities support different answers, which are rare in natural samples.
Masking or dropping one modality does not suffice, because a missing input does not contradict the other.
We instead pair samples of $\mathcal{D}$ that share a question but have different answers, and collect these pairs in $\mathcal{P}$.
Each pair $\{o, x\} \in \mathcal{P}$ yields a grid of four cells, one for each image--audio combination,
\begin{equation}
\label{eq:grid}
\mathcal{G}(o, x) = \bigl\{ (q, i_u, a_v) \;:\; u, v \in \{o, x\} \bigr\}.
\end{equation}
The two cells with $u = v$ are the original samples, and the two \emph{counterfactual} cells with $u \neq v$ combine the image of one sample with the audio of the other (Figure~\ref{fig:teaser}, right).
The grid contains only real media and both swap directions, so every image and audio clip of a pair appears in a matched and a mismatched cell.
Because the two samples have different answers, the image and the audio of the two counterfactual cells disagree, for example a cello on screen while a trumpet plays.

\subsection{Designated-Modality Relabeling}
\label{sec:relabel}

Each cell of the grid then needs a label.
A mismatched cell could be labeled as a mismatch, which trains a model to detect inconsistent media \citep{wvss}.
Such a label depends on whether the media match and not on the designated modality.
We instead assign each cell the answer of the sample that supplied its designated modality, so for a cell $c = (q, i_u, a_v)$ the label is
\begin{equation}
\label{eq:relabel}
y^{\star}(c) =
\begin{cases}
y_v & \text{if } d(q) = \{\text{audio}\}, \\
y_u & \text{if } d(q) = \{\text{image}\}, \\
y_u & \text{if } d(q) = \{\text{image}, \text{audio}\} \text{ and } u = v .
\end{cases}
\end{equation}
For a question designating both modalities, a counterfactual cell has no known label, so we retain only the two original cells.
Figure~\ref{fig:relabel} shows the labels for a pair with answers $g = y_o$ and $g' = y_x$.

With these labels, the designated input is the only input that predicts the label.
In a complete Audio or Visual grid, each answer labels two cells.
These two cells contain different non-designated inputs, and one is matched while the other is mismatched.
Formally, let $\mathrm{match}(c) = \mathbb{I}[u = v]$ indicate whether a cell is matched, and let $c_{\bar{d}}$ denote its non-designated input, which is the image for an Audio question and the audio for a Visual question.
For a cell $c$ drawn uniformly from the grid,
\begin{equation}
\label{eq:balance}
\Pr\bigl[y^{\star}(c) = g \mid c_{\bar{d}}\bigr] \;=\; \Pr\bigl[y^{\star}(c) = g \mid \mathrm{match}(c)\bigr] \;=\; \tfrac{1}{2}.
\end{equation}
Neither the non-designated input nor the match indicator alone predicts the label better than chance, whereas the designated input determines it through Equation~\ref{eq:relabel}.
In the original samples alone, the two inputs always occur together, so both predict the label equally well, which is the origin of the shortcut.
In joint training on Audio and Visual grids, the same image determines the label under a Visual question and is uninformative under an Audio question.
The model must therefore infer from the question which modality to use, rather than acquire a fixed preference.

\subsection{Answer-Token Supervision}
\label{sec:supervision}

Relabeling gives each counterfactual cell a target answer but no reference response, so standard supervised fine-tuning on complete responses is not applicable.
The pilot study also shows that RL rewards on the generated text do not remove the shortcut.
DMC-Repair therefore supervises the answer directly and trains the model to prefer the label of each cell over the other answer of its pair.
For a cell $c$ and a candidate answer $y = (y_1, \dots, y_{|y|})$, we teacher-force the prompt followed by \texttt{<answer>} and score the answer by its length-normalized log-likelihood,
\begin{equation}
\label{eq:score}
s_\theta(y \mid c) = \frac{1}{|y|} \sum_{t=1}^{|y|} \log p_\theta\bigl(y_t \mid c, \texttt{<answer>}, y_{<t}\bigr).
\end{equation}
Let $\bar{y}(c)$ be the other answer of the pair.
The loss of a cell is a hinge on the score gap with threshold~$\tau$, and DMC-Repair minimizes its average over the set $\mathcal{C}$ of all labeled cells,
\begin{equation}
\label{eq:hinge}
\mathcal{L}(\theta) = \frac{1}{|\mathcal{C}|} \sum_{c \in \mathcal{C}} \max\Bigl(0,\; \tau - \bigl[s_\theta(y^{\star}(c) \mid c) - s_\theta(\bar{y}(c) \mid c)\bigr]\Bigr).
\end{equation}
For one-token answers, the score gap is the answer margin of Section~\ref{sec:design} oriented toward the label, so the objective optimizes the quantity that the diagnostic measures.
The hinge is zero once this margin reaches $\tau$, so only cells whose margin remains below $\tau$ contribute gradients.
Since the loss covers only the answer tokens, it sets no target for the generated description or reasoning, and training requires no rollouts, reward model, or preference pairs.

\subsection{The Recipe in General}
\label{sec:general}

The construction extends beyond images and audio to any dataset that provides a designation $d(Q)$ for every question and real samples that share a question but differ in their answers.
With $N$ modality streams, we take each stream $k$ from a source sample $s_k$ to form a cell $c_{\mathbf{s}} = (Q, M_1^{(s_1)}, \dots, M_N^{(s_N)})$.
Its label is the answer of the sample that supplied all designated streams,
\begin{equation}
\label{eq:general}
y^{\star}(c_{\mathbf{s}}) = y_{s_k} \ \text{ for } k \in d(Q), \qquad \text{defined only if } s_k = s_{k'} \text{ for all } k, k' \in d(Q).
\end{equation}
The label depends only on the designated streams, as Equation~\ref{eq:invariance} requires.
We next measure how closely models trained on these labels meet this requirement.

\section{Experiments}
\label{sec:exp}

We organize the experiments around four questions.
\textbf{Q1:} Does DMC-Repair suppress the cross-modal shortcut on held-out questions and in another model family (Section~\ref{sec:main_results})?
\textbf{Q2:} Which parts of the grid construction produce the gain, and do the loss and parameterization matter (Section~\ref{sec:dissection})?
\textbf{Q3:} Does the suppression generalize zero-shot to unseen data, and how does it compare with published methods (Section~\ref{sec:transfer})?
\textbf{Q4:} Can a modality reward in RL achieve the same effect, and does the suppression persist through subsequent RL without degrading other abilities (Section~\ref{sec:analysis})?

\subsection{Datasets and Metrics}
\label{sec:setup_exp}

\textbf{Datasets.}
Before any measurement, we split the 387 MUSIC-AVQA diagnostic families (Appendix~\ref{app:details}) into a development set of 130 families (65 Audio, 27 Visual, and 38 Audio-Visual), which serves as the validation set to evaluate all design choices, and a confirmation set of 257 families, which serves as the test set and remains unseen until all checkpoints are frozen.
After excluding every diagnostic family and its media, we build 6{,}176 training cells from 2{,}036 MUSIC-AVQA pairs in 643 question families.
AVQA \citep{avqa} with 104 families and AVHBench \citep{avhbench} with 5{,}302 questions are used to evaluate the zero-shot generalization ability, and IntentBench \citep{humanomniv2} with 2{,}689 questions is used to evaluate the general omni-modal reasoning ability.

\textbf{Metrics.}
Under direct scoring, we report the \emph{Shortcut Index} $\si = |\ev| / (|\ea| + |\ev|)$ of Section~\ref{sec:design}, the image's share of the answer effect, which is ideally zero on Audio questions and one on Visual questions.
We also report the non-designated effect, which is $\ev$ on Audio questions and $\ea$ on Visual questions.
Under free generation, we report \emph{Audio-Following}, the rate at which answers on conflict cells follow the audio, and \emph{Matched-Cell Accuracy}.
Unless stated otherwise, intervals in brackets are 95\% family-cluster bootstrap confidence intervals (CIs) over 10{,}000 resamples, paired between checkpoints.
Appendices~\ref{app:details} and~\ref{app:prompts} give the training and statistical details and all prompts.

\subsection{Main Results}
\label{sec:main_results}

\begin{table}[t]
\caption{Repair results on held-out development families.
The confirmation-set row uses the 128 Audio families of the sealed confirmation set.
Full and LoRA \citep{lora} both start from SFT$_{\mathrm{clean}}$ and use the same cells (Appendix~\ref{app:details}). Full (bold) updates all LLM parameters, and LoRA trains a rank-16 adapter. The released HumanOmniV2 checkpoint \citep{humanomniv2} is evaluated by direct scoring only.
$\Delta$ is the change of Full from SFT$_{\mathrm{clean}}$ with its paired 95\% confidence interval (CI).
In the own-reasoning block, $\si$ is rescored after the model's own reasoning, and the other two rows use free generation with unparseable outputs counted as failures (Appendix~\ref{app:prompts}).}
\label{tab:main}
\centering
\footnotesize
\setlength{\tabcolsep}{2.5pt}
\begin{tabular}{llccccl@{\hspace{4pt}}l}
\toprule
 & & HumanOmniV2 & SFT$_{\mathrm{clean}}$ & LoRA & Full & \multicolumn{2}{l}{$\Delta$ (Full) [95\% CI]} \\
\midrule
\multirow{3}{*}{\begin{tabular}[c]{@{}l@{}}Audio q.,\\direct\end{tabular}} & $\si$ (ideal $=0$) & 0.498 & 0.549 & 0.194 & \textbf{0.187} & $-0.362$ & [$-0.43$, $-0.29$] \\
 & $\ev$ (non-designated) & $+1.40$ & $+1.05$ & $-0.04$ & $\mathbf{+0.07}$ & $-0.98$ & [$-1.39$, $-0.59$] \\
\cmidrule(l){2-8}
 & $\si$, confirmation set & --- & 0.559 & 0.208 & \textbf{0.224} & $-0.335$ & [$-0.39$, $-0.28$] \\
\midrule
\multirow{2}{*}{\begin{tabular}[c]{@{}l@{}}Visual q.,\\direct\end{tabular}} & $\si$ (ideal $=1$) & 0.716 & 0.713 & 0.872 & \textbf{0.872} & $+0.159$ & [$+0.09$, $+0.22$] \\
 & $\ea$ (non-designated) & $+1.86$ & $+1.28$ & $-0.10$ & $\mathbf{+0.02}$ & $-1.26$ & [$-1.65$, $-0.87$] \\
\midrule
\multirow{3}{*}{\begin{tabular}[c]{@{}l@{}}Audio q.,\\own\\reasoning\end{tabular}} & $\si$ (ideal $=0$) & --- & 0.612 & 0.256 & \textbf{0.271} & $-0.341$ & [$-0.42$, $-0.27$] \\
 & Audio-Following, \% & --- & 43.1 & 60.8 & \textbf{61.2} & $+18.1$ & [$+10.0$, $+26.2$] \\
 & Matched-Cell Accuracy, \% & --- & 61.5 & 60.0 & \textbf{67.3} & $+5.8$ & [$-0.4$, $+11.9$] \\
\bottomrule
\end{tabular}
\end{table}

\textbf{Results on MUSIC-AVQA.}
DMC-Repair shifts answers to each question's designated modality (Table~\ref{tab:main}).
On Audio questions, it reduces the Shortcut Index by two thirds, whereas the released HumanOmniV2 remains near one half.
The repair lowers the image effect for either audio clip and brings it close to zero (Appendix~\ref{app:interaction}).
On Visual questions, the index moves toward its ideal value of one, and the audio effect also falls close to zero.
The model thus learns which modality each question requires rather than suppressing one.
The index reduction persists when answers are rescored after the model's own reasoning.
In free generation, 84 of the 260 conflict-cell answers switch toward the audio and 37 away from it.
Matched-cell accuracy on Audio questions rises as well (Table~\ref{tab:main} and Appendix~\ref{app:qtype}).
The gain extends to generated answers without reducing Audio-question accuracy.

\textbf{Results on the confirmation set.}
On the sealed confirmation set, the repair reduces the Shortcut Index by $59.9\%$ (Table~\ref{tab:main}) and improves audio-following by $13.7$ points.
All seven pre-registered tests meet their criteria (Appendix~\ref{app:conf}), so the gain is not an artifact of tuning on the development set.

\textbf{Cross-model generalization.}
The repair is also effective on MiniCPM-o-2.6 \citep{minicpmv,minicpmo}, whose encoders and output format differ from Qwen2.5-Omni's.
LoRA training on the same cells reduces its Shortcut Index by about half on both sets under its own answer scoring and improves its zero-shot AVHBench accuracy by $1.4$ points [$+0.4$, $+2.3$] (Appendices~\ref{app:xmodel} and~\ref{app:avh}).
In both model families, DMC-Repair suppresses the cross-modal shortcut on held-out questions.

\subsection{Ablation Study}
\label{sec:dissection}

\begin{figure}[t]
\centering
\includegraphics[width=0.95\textwidth]{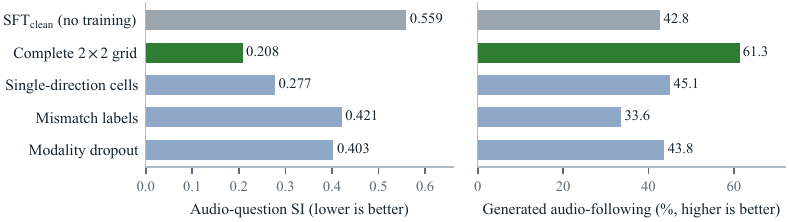}
\caption{Construction controls on the confirmation set (128 Audio families, 510 conflict cells).
The complete grid is the LoRA reference, and paired differences with 95\% CIs are in Appendix~\ref{app:conf}.}
\label{fig:controls}
\end{figure}

\textbf{Construction controls.}
We next ask which elements of the construction produce the gain.
Three pre-registered controls each remove one element (Figure~\ref{fig:controls}) and show that the gain requires both conflicting inputs and designated-modality labels.
Modality dropout retains the labels but removes the conflicts, and mismatch labels retain the conflicts but teach the model to report a mismatch.
Both leave the index at about twice that of the complete grid.
Single-direction cells cover one swap direction per pair and retain about $80\%$ of the index reduction.
In free generation, the complete grid leads every control by $12$ to $31$ points under three treatments of unparseable outputs (Appendix~\ref{app:conf}).
On the development set, training only on the original cells yields about a third of the index reduction without improving audio-following (Appendix~\ref{app:ablation}).
Every element of the construction matters.

\textbf{Loss and parameterization.}
The loss and parameterization matter much less.
Cross-entropy achieves $89\%$ of the reduction obtained with the hinge loss, and LoRA matches full fine-tuning with $148$ times fewer trainable parameters.
The index falls by $0.30$ to $0.36$ for all four seeds (Appendix~\ref{app:ablation}).
The gain originates from the training cells rather than the training settings.

\subsection{Zero-Shot Generalization Performance}
\label{sec:transfer}

Since all training cells are drawn from MUSIC-AVQA, we next apply the repaired model directly to an unseen dataset and an unseen benchmark.
In this zero-shot setting, the checkpoints are frozen before evaluation, and neither target is used for training, model selection, or prompt tuning.

\begin{table}[t]
\caption{Comparison with published methods on AVHBench, all built on Qwen2.5-Omni.
Base is the audio-hallucination accuracy that each paper reports for its base model, and each change is measured against that base model on the same task.
Other numbers are copied from the papers (---, not reported), and Table~\ref{tab:avh} gives all tasks for our models.
$^{\dagger}$Evaluated by \citet{mad}.}
\label{tab:avh_compare}
\centering
\footnotesize
\setlength{\tabcolsep}{2.5pt}
\begin{tabular}{llccc}
\toprule
Method & Training data & Base & $\Delta$ Audio hall. & $\Delta$ AV matching \\
\midrule
AVCD$^{\dagger}$ \citep{avcd} & none (decoding) & 73.0 & $+2.8$ & --- \\
MAD \citep{mad} & none (decoding) & 73.0 & $+5.7$ & --- \\
\midrule
ACPO \citep{acpo} & audio-swap preferences & 66.7 & $+2.6$ & --- \\
OmniDPO \citep{omnidpo} & audio-visual preferences & 67.6 & $+9.9$ & --- \\
MoD-DPO++ \citep{moddpo} & modality-perturbed preferences & 77.4 & $+6.0$ & $+15.0$ \\
\citet{somemodalities} & (mis)aligned AudioSet clips & 71.7 & $+8.2$ & $+0.0$ \\
\midrule
DMC-Repair (full) & MUSIC-AVQA grid cells & 61.9 & $+5.8$ & $+7.2$ \\
\quad + subsequent RL & \quad + GRPO & 61.9 & $\mathbf{+11.3}$ & $+7.3$ \\
\bottomrule
\end{tabular}
\end{table}

\textbf{Generalization to an unseen dataset.}
DMC-Repair generalizes zero-shot to AVQA \citep{avqa}, an everyday audio-visual QA dataset built on VGGSound \citep{vggsound}.
The selected questions, taken from the AVQA items of OmniInstruct \citep{omnibench}, ask whether an object is the main sound source.
On these questions, all three pre-registered criteria are met (Appendix~\ref{app:avqa}).
The repair reduces the image effect by $62\%$ and improves audio-following by $18$ points (as on \mbox{MUSIC-AVQA}), and matched-cell accuracy rises by $14$ points.
This suggests that the model learned a rule for selecting the modality rather than knowledge about particular instruments or scenes.

\textbf{Generalization to a hallucination benchmark.}
Suppressing the shortcut also improves zero-shot accuracy on the audio-hallucination and audio-visual matching tasks of AVHBench \citep{avhbench}, the two tasks whose answers depend on the audio.
Video-hallucination accuracy is nearly unchanged (Appendix~\ref{app:avh}).
Without any hallucination-specific training data, the repair achieves an audio-hallucination gain over its base that is comparable to that of the strongest published decoding method.
Subsequent RL nearly doubles this gain, which is then the largest in Table~\ref{tab:avh_compare}.
Most of the audio gain is due to audible objects that the starting checkpoint missed.
Accuracy on audible objects rises from $49.1\%$ to $61.7\%$ after repair and to $75.4\%$ after RL.
The suppression generalizes beyond the training data and improves the abilities that require listening.

\subsection{Analysis}
\label{sec:analysis}

\begin{figure}[t]
\centering
\includegraphics[width=\textwidth]{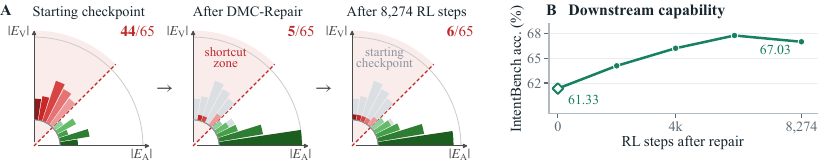}
\caption{Column area is proportional to the number of Audio families per $9^\circ$ angle of $(|\ea|,|\ev|)$.}
\label{fig:retention}
\end{figure}

\textbf{Modality rewards in RL.}
\phantomsection\label{sec:rewards_body}%
Adding a modality reward to RL in place of the repair does not remove the shortcut (Appendix~\ref{app:rewards}).
Four of our six RL variants use a judge to score the model's own description of the input, and the three judge rewards evaluated on all families increase the image effect.
GRPO with only format and accuracy rewards increases the image effect about twice as much as the audio effect.
Mixing grid cells into the RL data lowers the index by less than a tenth of the repair's effect.

\textbf{Robustness to subsequent RL.}
\phantomsection\label{sec:retention}%
When GRPO follows the repair, the suppression persists through all $8{,}274$ steps, over which IntentBench accuracy improves (Figure~\ref{fig:retention} and Appendix~\ref{app:conf}).
Of the 44 development Audio families in the shortcut zone ($\si > 0.5$) before the repair, 41 leave and 2 others enter, and 6 of the 65 lie in it at the RL endpoint.
On the confirmation set, the index is unchanged and audio-following rises by another $9$ points, which meets both pre-registered retention criteria.

\textbf{Other abilities.}
The repair itself has little effect on other abilities.
Before subsequent RL, \mbox{IntentBench} accuracy remains within one point of the starting checkpoint ($61.33\%$ versus $60.64\%$), and accuracy on Audio-Visual questions, for which we retain only the two original cells, does not decrease (Appendix~\ref{app:qtype}).
Overall, RL does not replace the repair but can follow it without undoing it.

\subsection{Case Study}
\label{sec:case}

\begin{figure}[t]
\centering
\includegraphics[width=\textwidth]{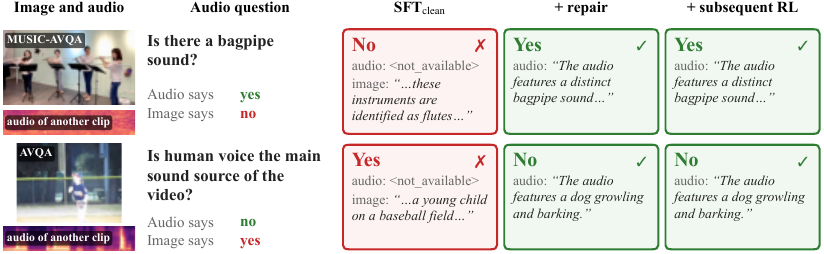}
\caption{Generated answers on two conflict cells, colored by the modality that they follow.}
\label{fig:qual}
\end{figure}

Finally, Figures~\ref{fig:teaser} and~\ref{fig:qual} show how the repair changes the generated text, and Appendix~\ref{app:qual} adds eight more cells, including two on which all three checkpoints follow the image.
The starting checkpoint leaves its audio description empty and bases its answer on the image.
In contrast, the repaired model and the RL endpoint describe the audio content, namely the bagpipe in row~1 of Figure~\ref{fig:qual} and the barking dog in row~2 from the unseen AVQA, and base their answers on the audio.

\section{Conclusion}
\label{sec:conclusion}

Omni-modal LLMs should answer audio questions from the audio, but their training data and rewards rarely separate the image from the audio.
By swapping the two between real samples, our Factorized Modality Diagnostic shows that the tested models rely on the image about as much as on the audio, at every stage of SFT and RL.
To suppress this cross-modal shortcut, \mbox{DMC-Repair} recombines real samples into grids, labels each cell by its designated modality, and supervises only the answer tokens.
The repair reduces the Shortcut Index by $59.9\%$ on a sealed confirmation set, generalizes across two model families and zero-shot to an unseen dataset and benchmark, and persists through subsequent RL that improves task accuracy.
Future work can extend the relabeling rule of Section~\ref{sec:general} to questions that designate several modalities and to open-ended answers.

\subsection*{AI use statement}
We used generative AI coding and writing assistants to implement author-specified methods, clean and reformat datasets, review conceptual framing and experimental design, assess pre-registered hypotheses and criteria, and assist with result interpretation.
The one model-assisted change to training data was the split of the cold-start context annotation into separate audio and visual descriptions, where the model could only assign sentences of the original annotation and a coverage check rejected any split that dropped content.
All research decisions were made by the authors.
We did not use generative AI to synthesize datasets or labels or to formulate or prove mathematical claims.
Translation and qualitative or thematic data analysis are not applicable to this work.

AI tools also assisted with drafting and editing the paper, drawing and editing figures, creating and editing code, finding and summarizing related work, and formatting references.
The authors reviewed all AI-assisted work.
Code was reviewed and tested, analysis scripts were checked against independent computations, and reported numbers were recomputed from raw outputs.
Confirmation results were computed with the frozen analysis script described in Appendix~\ref{app:conf}.
Text was checked against the results and citations against the original publications.
The authors take responsibility for the final text, claims, and artifacts.

\subsection*{Ethics statement}
The study uses publicly released benchmarks and involves no new human-subject data collection. MUSIC-AVQA is used under its release license.
We are not aware of ethical concerns specific to this evaluation and training-data construction study.

\subsection*{Reproducibility statement}
Section~\ref{sec:design} defines the diagnostic and its statistics, and Section~\ref{sec:method} specifies how training cells are built and supervised.
Appendix~\ref{app:details} gives training and evaluation settings, and Appendix~\ref{app:conf} gives the pre-registered confirmation protocol and its criteria.
The diagnostic sets use MUSIC-AVQA and AVQA, the training cells are built from MUSIC-AVQA, and the external evaluations use IntentBench and AVHBench, all of which are publicly released.
The core scripts for the diagnostic, cell construction, training, and analysis are available at the anonymous repository linked in the abstract, together with the family lists of the diagnostic sets, and the complete code will be released upon acceptance.

\bibliography{refs_iclr}
\bibliographystyle{iclr2027_conference}

\clearpage
\appendix
\raggedbottom
\section*{Overview of the Appendices}
\begin{itemize}[nosep, leftmargin=*]
\item \textbf{Setup.} Appendix~\ref{app:details} gives the training and evaluation details, and Appendix~\ref{app:prompts} lists the prompts.
\item \textbf{Diagnostic.} Appendix~\ref{app:audit_full} extends the pilot study to every configuration, Appendix~\ref{app:interaction} examines the interaction term, and Appendix~\ref{app:qtype} breaks the results down by question type.
\item \textbf{Repair.} Appendix~\ref{app:rewards} shows that none of the six RL variants of Section~\ref{sec:rewards_body} removes the shortcut, Appendix~\ref{app:conf} reports the pre-registered confirmation tests, Appendix~\ref{app:ablation} ablates the loss and the parameterization, and Appendix~\ref{app:xmodel} gives the cross-model diagnostic on MiniCPM-o-2.6.
\item \textbf{Zero-shot generalization.} Appendices~\ref{app:avqa} and~\ref{app:avh} give the AVQA and AVHBench protocols.
\item \textbf{Case study and related work.} Appendix~\ref{app:qual} adds eight conflict cells to the case study, and the related work continues in Appendix~\ref{app:related}.
\end{itemize}

\section{Training and Evaluation Details}
\label{app:details}

This appendix gives the settings behind Sections~\ref{sec:si_table} and~\ref{sec:exp} and traces how the two diagnostic sets are built.

\textbf{SFT.}
SFT$_{\mathrm{clean}}$ is obtained by training the LLM backbone for two epochs with a learning rate of $2\times10^{-5}$, a batch size of 1 per GPU, and no gradient accumulation.
The vision and audio encoders remain frozen.
All training runs use four NVIDIA A800-SXM4-80GB GPUs.

\textbf{HumanOmniV2.}
We adopt HumanOmniV2 \citep{humanomniv2} as the baseline model because it is an open omni-modal reasoning model obtained from Qwen2.5-Omni by SFT and GRPO post-training, and our starting checkpoint is trained on its cold-start data.

\textbf{Repair recipe.}
The fully fine-tuned model (Full) is the repaired model used for confirmation and subsequent RL, and the LoRA model is the reference for the construction controls.
The training pool is built from MUSIC-AVQA after every question family and every media file of the diagnostic sets has been removed, so that no training cell shares content with the evaluation.
It comprises 461 Audio pairs and 591 Visual pairs with four cells each and 984 Audio-Visual pairs with two cells each (Section~\ref{sec:relabel}), and its answers take 42 distinct values.
In the LoRA recipe, rank-16 adapters ($\alpha{=}32$, dropout $0.05$) are trained on the LLM and both encoders for three epochs over the 6{,}176-cell pool, which takes about four hours.
This recipe uses a learning rate of $1\times10^{-4}$, gradient accumulation over 4 steps, and bf16 precision.
The hinge threshold is $\tau = 1.0$ over length-normalized candidate scores, and the base weights of the vision and audio encoders remain frozen.
Full fine-tuning follows the same schedule with a learning rate of $1\times10^{-5}$ and AdamW with bf16 optimizer states, and it takes about three hours.
The LoRA adapters are merged before evaluation.

\textbf{Construction controls.}
The controls use the LoRA recipe of the complete-pool reference, with changes to the training cells described in Section~\ref{sec:dissection}.
The single-direction control contains 5{,}124 cells and is trained for 4 epochs, which roughly matches the number of cells that the complete pool provides in 3 epochs.
The other two controls each use 6{,}176 cells for 3 epochs.

\textbf{Subsequent RL.}
GRPO training starts from the repaired model and runs for the configured budget of 8{,}274 steps on the RL training data of HumanOmniV2, filtered as described below.
It uses 2 rollouts per prompt, a learning rate of $5\times10^{-6}$, a maximum sequence length of 2{,}048, and gradient checkpointing.
As in HumanOmniV2, the KL coefficient decays linearly from $0.04$ to $0.01$ over the first half of training.
Rollout filtering retains prompts with accuracy in $(0, 0.75]$.

\textbf{Cross-model adaptation.}
MiniCPM-o-2.6 is evaluated with a neutral prompt, and its teacher-forced yes/no margin is read at the first generated token position as $m = \log p(\texttt{yes}) - \log p(\texttt{no})$.
The LoRA adapters are injected only into the LLM through the \texttt{inject\_adapter\_in\_model} interface of the PEFT library with the same rank, $\alpha$, and schedule as the Qwen recipe.

\textbf{Statistics.}
Confidence intervals are 95\% family-cluster bootstrap intervals over 10{,}000 resamples unless stated otherwise.

\textbf{Diagnostic set construction.}
Table~\ref{tab:index} traces both diagnostic sets from the source annotations to the retained families.
Among the candidate families, almost all removals result from the requirement of at least two samples for each answer, and the media-uniqueness constraint of Appendix~\ref{app:avqa} removes only one AVQA family.

\begin{table}[t]
\caption{Construction of the two diagnostic sets, from source annotations to retained families.
The first two data rows count annotation rows, and the next five rows count families.
Samples that share a question form a family, and we retain only families with at least two \texttt{yes} and two \texttt{no} samples.
One Audio pair of the MUSIC-AVQA confirmation set was later dropped for missing media, giving 773 pairs rather than 774. The media-uniqueness constraint of Appendix~\ref{app:avqa} leaves six AVQA families with a single pair, giving 202 pairs rather than 208.}
\label{tab:index}
\centering
\footnotesize
\setlength{\tabcolsep}{4pt}
\begin{tabular}{lrr}
\toprule
 & MUSIC-AVQA & AVQA \\
\midrule
Source annotation rows & 42{,}492 & 6{,}402 \\
\quad usable answer format & 19{,}478 & 2{,}343 \\
\midrule
Candidate families & 2{,}521 & 846 \\
\quad $-$ fewer than two \texttt{yes} or two \texttt{no} & $-2{,}134$ & $-741$ \\
\quad $-$ media already claimed by another family & --- & $-1$ \\
\quad $-$ overlapping the frozen evaluation manifest & $-0$ & --- \\
\midrule
Retained families & \textbf{387} & \textbf{104} \\
Pairs / cells & 773 / 3{,}092 & 202 / 808 \\
Question types & 193 Audio, 80 Visual, 114 A-V & 104 Audio \\
\bottomrule
\end{tabular}
\end{table}

\section{The Diagnostic Across Model Families and Prompt Formats}
\label{app:audit_full}

\begin{figure}[!b]
\centering
\includegraphics[width=0.66\textwidth]{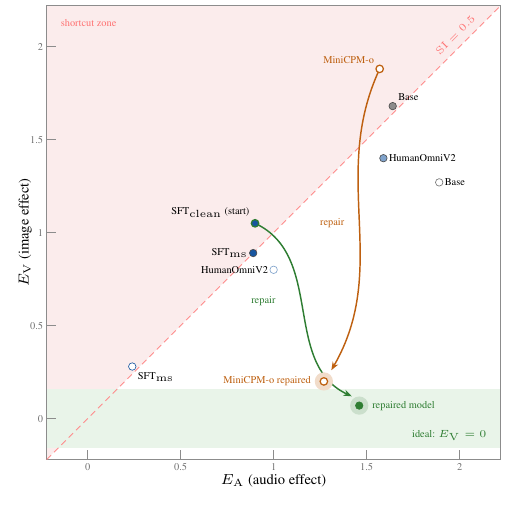}
\caption{The diagnostic of Figure~\ref{fig:audit} with all seven configurations of Table~\ref{tab:si} and both repairs.
Filled markers are native-prompt runs and open markers neutral-prompt runs.
The MiniCPM-o arrow starts at its base model, since no SFT stage precedes it.}
\label{fig:audit_full}
\end{figure}

This appendix shows that the shortcut appears in every tested configuration and that the diagnostic measures modality use (Section~\ref{sec:si_table}).
Figure~\ref{fig:audit} shows only the Qwen checkpoints under the native prompt, so that the change produced by the repair is easy to follow.
Figure~\ref{fig:audit_full} adds the neutral-prompt configurations and MiniCPM-o-2.6, whose repair is the exploratory cross-model generalization experiment of Section~\ref{sec:main_results} (Appendix~\ref{app:xmodel}).
Table~\ref{tab:si} lists the pilot-study diagnostic for every configuration other than SFT$_{\mathrm{clean}}$, which Table~\ref{tab:main} reports.
Every baseline configuration lies near the $\si = 0.5$ diagonal regardless of model family or prompt, and both repairs move toward $\ev = 0$.
Under the native prompt, HumanOmniV2 lies within the paired equivalence margin $|\Delta\si| < 0.1$ of its base model (Table~\ref{tab:si}).
The shortcut thus predates the post-training of HumanOmniV2 and persists through it.

\begin{table}[t]
\caption{Audio-question diagnostic on 65 development families, where the ideal $\si$ is $0$. Base is Qwen2.5-Omni, HumanOmniV2 is the released checkpoint of \citet{humanomniv2}, and SFT$_{\mathrm{ms}}$ is an earlier version of SFT$_{\mathrm{clean}}$. The neutral prompt omits the system prompt and output format. The paired native-prompt change in $\si$ from Base to HumanOmniV2 is $+0.021$ [$-0.019$, $+0.061$].}
\label{tab:si}
\centering
\small
\begin{tabular}{llccc}
\toprule
Model & Prompt & $\ea$ & $\ev$ & $\si$ [95\% CI] \\
\midrule
Base & native  & +1.64 & +1.68 & 0.478 [0.416, 0.538] \\
Base & neutral & +1.89 & +1.27 & 0.419 [0.366, 0.475] \\
HumanOmniV2 & native  & +1.59 & +1.40 & 0.498 [0.441, 0.553] \\
HumanOmniV2 & neutral & +1.00 & +0.80 & 0.486 [0.431, 0.539] \\
SFT$_{\mathrm{ms}}$ & native  & +0.89 & +0.89 & 0.484 [0.424, 0.543] \\
SFT$_{\mathrm{ms}}$ & neutral & +0.24 & +0.28 & 0.565 [0.504, 0.621] \\
MiniCPM-o-2.6 & neutral & +1.57 & +1.88 & 0.554 [0.501, 0.608] \\
\bottomrule
\end{tabular}
\end{table}

\textbf{Validity checks.}
Several checks show that the diagnostic measures modality use.
Shuffling the family-to-media mapping 1{,}000 times centers $\ea$ and $\ev$ on zero, with the observed effects outside the permutation nulls ($p<0.001$), so the effects require media that carry the family's answers.
Answer-level readouts agree with the index.
Swaps pull answers toward the incoming sample's ground truth at similar rates for audio and images ($25.0\%$ versus $21.2\%$ for the base model).
On conflict cells HumanOmniV2 follows the audio in exactly half of the direct answer comparisons, matching its $\si \approx 0.5$.
On Visual questions, the diagnostic gives $\ev > 2\ea$ (Appendix~\ref{app:qtype}), which rules out an index that returns one half by construction.
Total sensitivity $|\ea| + |\ev|$ varies more than sixfold across the configurations of Table~\ref{tab:si} while $\si$ remains nearly constant, so the index does not simply track the size of the margins.

\section{Modality Reward Designs}
\label{app:rewards}

This appendix details the six RL variants of Section~\ref{sec:rewards_body} and shows that none of them removes the cross-modal shortcut.
Every run retains the standard format and accuracy rewards and comprises 1{,}000 GRPO steps on rollouts restricted to prompts of intermediate difficulty.
The four judge designs add a modality reward.
In each judge design, a judge scores the response's own description of the input, and the score is added to the reward with weight $w$.
The four judge runs start from SFT$_{\mathrm{ms}}$, and the two runs without a modality judge start from SFT$_{\mathrm{clean}}$.
The two checkpoints share their training samples and hyperparameters, and both are trained on cold-start data whose description is split into separate audio and visual parts so that a judge can score each modality separately.
SFT$_{\mathrm{ms}}$ uses the first version of this split and SFT$_{\mathrm{clean}}$ a revised version.

The four judge designs differ in how the description is scored.
One design (graded description score) uses an external model to score the description on an open scale.
In two others, a frozen judge model answers the question from the description alone, with the media hidden.
One of them (description gain) bases the reward on how much the description improves the judge's answer.
The other (description and answer) combines the correctness of the judge's answer with the correctness of the answer in the response.
In the last (yes/no description score), a model that hears the audio or sees the frames returns a yes/no verdict on whether the description is grounded and sufficient.
Each change in Table~\ref{tab:interventions} is measured against that run's own starting checkpoint.

\begin{table}[!ht]
\caption{Changes in the Audio-question diagnostic after RL, paired against each run's own starting checkpoint (65 families unless noted).
The Start column gives the starting checkpoint (ms for SFT$_{\mathrm{ms}}$, clean for SFT$_{\mathrm{clean}}$), and $w$ is the weight of the modality reward.
Positive $\Delta\ev$ indicates a larger image effect. Bold marks the design that increases only the image effect.}
\label{tab:interventions}
\centering
\footnotesize
\setlength{\tabcolsep}{2.5pt}
\resizebox{\linewidth}{!}{%
\begin{tabular}{llccc}
\toprule
Reward design & Start & $\Delta\ea$ [95\% CI] & $\Delta\ev$ [95\% CI] & $\Delta\si$ [95\% CI] \\
\midrule
Graded description score ($w{=}1.0$) & ms & $-0.012$ [$-0.03$, $+0.01$] & $\mathbf{+0.053}$ [$+0.03$, $+0.08$] & $+0.019$ [$+0.005$, $+0.033$] \\
Description gain ($w{=}0.3$) & ms & $+0.051$ [$+0.03$, $+0.07$] & $+0.069$ [$+0.04$, $+0.09$] & $-0.002$ [$-0.018$, $+0.015$] \\
Description and answer ($w{=}0.3$) & ms & $+0.028$ [$+0.02$, $+0.04$] & $+0.035$ [$+0.02$, $+0.05$] & $+0.005$ [$-0.010$, $+0.020$] \\
Yes/no description score ($w{=}0.3$) & ms & $-0.029$ [$-0.28$, $+0.22$] & $-0.178$ [$-0.40$, $+0.04$] & inconclusive (31 fam.) \\
\midrule
GRPO, no modality reward & clean & $+0.038$ [$+0.02$, $+0.06$] & $+0.074$ [$+0.05$, $+0.10$] & $+0.003$ [$-0.011$, $+0.016$] \\
GRPO + grid cells & clean & $+0.104$ [$+0.07$, $+0.14$] & $+0.058$ [$+0.03$, $+0.09$] & $-0.021$ [$-0.036$, $-0.007$] \\
\bottomrule
\end{tabular}}
\end{table}

The three judge rewards evaluated on all 65 families increase the image effect.
The graded score increases the image effect while leaving the audio effect unchanged.
Although its visual and audio scores enter the reward with equal weight, this judge awards the visual score roughly three times as often as the audio score, so the reward favors visual descriptions.
The description gain and the combined description-and-answer reward increase both effects, with a larger point estimate for images.
The yes/no design was evaluated on 31 families and is inconclusive.
Without a modality reward, GRPO increases the image effect about twice as much as the audio effect.
Mixing grid cells into the RL data lowers the index by less than a tenth of the repair's effect in Table~\ref{tab:main}.

\textbf{Swap-margin reward.}
We also screened a swap-margin reward but did not train it, because it would be zero for every completion.
The reward is granted only when the answer margin changes sign after the designated modality is swapped and keeps its sign after the non-designated modality is swapped.
Before training, we checked whether this reward varies across completions.
Across 179 prompts with four completions each, the margins obtained by rescoring the same completion under different media are almost perfectly correlated, and no completion satisfies the gating conditions.
After a completion, the answer margin depends far more on the completion text than on the media.

\section{Interaction Terms}
\label{app:interaction}

This appendix examines the interaction term of Section~\ref{sec:design} and shows that the repair lowers the image effect for either audio clip (Section~\ref{sec:main_results}).
The interaction $\psi$ measures how the effect of swapping one modality changes with the source of the other modality.
The sign of $\psi$ is defined only once the two items of a pair are ordered, so we use the orientation fixed in Section~\ref{sec:design}, in which the own item is always the one whose ground-truth answer is \texttt{yes}.
Under that convention, signed $\psi$ values are averaged over a family's two pairs and then across families, exactly like the main effects.
A negative $\psi$ means that swapping the audio has a smaller signed effect when the image is the own item's than when it is the partner's.
Equivalently, the mean margin of the two mismatched cells lies above that of the two matched cells.
Every configuration in Table~\ref{tab:interaction} shows this pattern, as expected if the margin saturates when both modalities support \texttt{yes}.
Since $|\psi|$ is smaller than both main effects of Table~\ref{tab:si} in every configuration, swapping either modality moves the mean margin in the same direction at both levels of the other modality.

\begin{table}[!ht]
\caption{Interaction terms $\psi$ for Audio questions, under the orientation and aggregation stated above.}
\label{tab:interaction}
\centering
\small
\begin{tabular}{lc}
\toprule
Configuration & $\psi$ [95\% CI] \\
\midrule
Base, native & $-0.482$ [$-0.718$, $-0.267$] \\
Base, neutral & $-0.556$ [$-0.813$, $-0.327$] \\
HumanOmniV2, native            & $-0.513$ [$-0.769$, $-0.277$] \\
HumanOmniV2, neutral           & $-0.353$ [$-0.511$, $-0.201$] \\
SFT$_{\mathrm{ms}}$, native    & $-0.359$ [$-0.505$, $-0.229$] \\
SFT$_{\mathrm{ms}}$, neutral   & $-0.052$ [$-0.113$, $+0.005$] \\
MiniCPM-o-2.6, neutral         & $-0.445$ [$-0.693$, $-0.218$] \\
\bottomrule
\end{tabular}
\end{table}

\subsection*{Conditional Image Effects}
\label{app:conditional}

Equation~\ref{eq:invariance} is a statement about each condition separately, so we report the image effect at each level of the audio as well as its average.
Writing the image effect at each level of the audio as
$\delta_{\mathrm{V}}(o) = m_{oo} - m_{xo}$ and $\delta_{\mathrm{V}}(x) = m_{ox} - m_{xx}$,
we have $\ev = \tfrac{1}{2}[\delta_{\mathrm{V}}(o) + \delta_{\mathrm{V}}(x)]$ and $\delta_{\mathrm{V}}(o) - \delta_{\mathrm{V}}(x) = 2\psi$, so $\psi$ measures how far the two conditional effects differ.
Table~\ref{tab:conditional} reports both conditional effects and $\psi$ for the starting checkpoint, the repaired model, and the RL endpoint, together with the mean over families of the absolute image effect $|\ev|$ of each family.

\begin{table}[!ht]
\caption{Image effect on Audio questions decomposed by the audio held fixed, 65 development families.
Own audio comes from the item whose answer is \texttt{yes}, and partner audio from the item whose answer is \texttt{no}.
The last two rows average over families the absolute image effect and the larger absolute conditional effect of each family, so effects of opposite sign cannot cancel. Intervals use 8{,}000 resamples.}
\label{tab:conditional}
\centering
\footnotesize
\setlength{\tabcolsep}{3pt}
\begin{tabular}{lccc}
\toprule
Statistic & SFT$_{\mathrm{clean}}$ & Repaired (full) & RL endpoint \\
\midrule
$\delta_{\mathrm{V}}(o)$, own audio & $+0.682$ [$+0.414$, $+0.966$] & $-0.027$ [$-0.149$, $+0.083$] & $+0.019$ [$-0.113$, $+0.146$] \\
$\delta_{\mathrm{V}}(x)$, partner audio & $+1.417$ [$+0.905$, $+1.970$] & $+0.165$ [$+0.016$, $+0.319$] & $+0.449$ [$+0.228$, $+0.684$] \\
$\psi$ & $-0.368$ [$-0.521$, $-0.233$] & $-0.096$ [$-0.181$, $-0.015$] & $-0.215$ [$-0.338$, $-0.096$] \\
\midrule
mean $|\ev|$ & $1.275$ [$0.939$, $1.651$] & $0.343$ [$0.278$, $0.413$] & $0.506$ [$0.417$, $0.602$] \\
mean $\max|\delta_{\mathrm{V}}|$ & $1.687$ [$1.220$, $2.191$] & $0.629$ [$0.524$, $0.742$] & $0.928$ [$0.775$, $1.100$] \\
\bottomrule
\end{tabular}
\end{table}

Both conditional effects are positive before the repair and decrease afterward, and the confidence interval of the own-audio effect then includes zero.
In the last row of Table~\ref{tab:conditional}, we take the larger absolute value of the two conditional image effects in each family and average these values across families.
The repair reduces this mean by more than half, so it lowers the image effect within families, not only on average.
At the RL endpoint, the partner-audio effect and both per-family means remain well below their pre-repair values.

\section{Per-Question-Type Breakdown}
\label{app:qtype}

This appendix breaks the development results down by question type to support the validity check of Appendix~\ref{app:audit_full} and the accuracy statements of Sections~\ref{sec:main_results} and~\ref{sec:analysis}.
On Visual questions, where the image is designated, every model in Table~\ref{tab:qtype} gives an image effect more than twice its audio effect.

\begin{table}[!ht]
\caption{Diagnostic by question type on the development set.
For Audio questions, the image is non-designated ($\si$ ideal $= 0$).
For Visual questions, the audio is non-designated ($\si$ ideal $= 1$).}
\label{tab:qtype}
\centering
\small
\begin{tabular}{llccc}
\toprule
Model & Q-type & $\ea$ & $\ev$ & $\si$ \\
\midrule
\multirow{2}{*}{Base} & Audio (65) & +1.635 & +1.682 & 0.478 \\
 & Visual (27) & +1.565 & +5.014 & 0.747 \\
\midrule
\multirow{2}{*}{HumanOmniV2} & Audio (65) & +1.591 & +1.403 & 0.498 \\
 & Visual (27) & +1.856 & +4.744 & 0.716 \\
\midrule
\multirow{2}{*}{MiniCPM-o-2.6} & Audio (65) & +1.565 & +1.883 & 0.554 \\
 & Visual (27) & +1.363 & +7.104 & 0.802 \\
\bottomrule
\end{tabular}
\end{table}

The repair improves matched-cell accuracy on Audio questions, and the changes on the other two question types have intervals that include zero (Table~\ref{tab:qtype_acc}).
A matched cell carries the media of a single real sample, so its ground-truth answer is that sample's own answer.
Accuracy is the rate at which this answer receives the higher score under direct scoring, averaged within a family and then across families.
Audio-Visual questions contribute only their two original cells to training (Section~\ref{sec:relabel}), yet their accuracy does not fall after repair and rises after subsequent RL.
Because matched cells are the original samples, the repair also improves Audio-question answers on unmodified inputs.

\begin{table}[!ht]
\caption{Matched-cell accuracy by question type on the development set, under the direct answer scoring of Section~\ref{sec:design}.
Values are percentages, and the final column gives the repaired model's paired change against SFT$_{\mathrm{clean}}$ with 95\% family-cluster bootstrap intervals over 8{,}000 resamples.}
\label{tab:qtype_acc}
\centering
\small
\begin{tabular}{lcccc}
\toprule
Q-type & SFT$_{\mathrm{clean}}$ & Repaired (full) & RL endpoint & $\Delta$ (full) \\
\midrule
Audio (65)        & 64.2 & 72.3 & 73.8 & $+8.1$ [$+2.3$, $+13.5$] \\
Visual (27)       & 83.3 & 80.6 & 81.5 & $-2.8$ [$-10.2$, $+4.6$] \\
Audio-Visual (38) & 50.7 & 53.9 & 61.8 & $+3.3$ [$-4.6$, $+11.2$] \\
\bottomrule
\end{tabular}
\end{table}

\section{Pre-Registered Confirmation Protocol}
\label{app:conf}

This appendix reports the pre-registered tests behind the held-out results of Sections~\ref{sec:main_results}, \ref{sec:dissection}, and~\ref{sec:retention}.
The confirmation set contains 257 families and was constructed together with the 130-family development set.
The confirmation set was held out from model selection, hyperparameter tuning, and analysis, and it was used once to evaluate the frozen checkpoints.
The seven hypotheses are reported in a pre-specified order, and if one of them does not meet its criterion, the subsequent results are reported descriptively.

\begin{table}[!ht]
\caption{Results for the pre-specified confirmation hypotheses on the 257-family confirmation set. All seven tests meet their criteria, and H2 is reported in two rows. The Audio-question $\si$ tests use 128 families, and H3 evaluates the 53 Visual families.
SFT, full, and RL denote the starting checkpoint, the repaired model, and the RL endpoint.
H4 uses the repair gain $B = \si_{\mathrm{SFT}} - \si_{\mathrm{full}}$ and the change through RL $D = \si_{\mathrm{RL}} - \si_{\mathrm{full}}$, observed as $0.000$ [$-0.024$, $+0.023$], and G2 uses the audio-following rates $p_S$, $p_F$, and $p_R$ of the three checkpoints.
H2 tests whether the repaired image effect lies within $\pm10\%$ of $|\ev(\mathrm{SFT})|$ around zero, and the interval of its upper row must lie below zero and that of its lower row above zero.
The lower block compares each construction control with the complete-pool LoRA reference as control minus reference, and its last column states whether the result goes in the pre-registered direction.}
\label{tab:conf}
\centering
\small
\setlength{\tabcolsep}{5pt}
\begin{tabular}{llc}
\toprule
Test & Statistic & Criterion met \\
\midrule
H1 (primary): repair $\Delta\si$ & $-0.335$ [$-0.390$, $-0.279$] & Yes \\
G1: repair audio-following $\Delta$ (points) & $+13.7$ [$+7.4$, $+19.7$] & Yes \\
G2: RL endpoint generation retention (points) & $+11.9$ [$+8.1$, $+15.6$] & Yes \\
H4: SI retention through RL ($\ge 80\%$) & $D - 0.2B = -0.067$ [$-0.089$, $-0.046$] & Yes \\
H3: Visual-question $\Delta\si$ & $+0.176$ [$+0.110$, $+0.241$] & Yes \\
H5: LoRA variant $\Delta\si$ & $-0.352$ [$-0.407$, $-0.294$] & Yes \\
H2 (upper): $\ev(\mathrm{full}) - 0.1\,|\ev(\mathrm{SFT})|$ & $-0.088$ [$-0.158$, $-0.017$] & Yes \\
H2 (lower): $\ev(\mathrm{full}) + 0.1\,|\ev(\mathrm{SFT})|$ & $+0.154$ [$+0.061$, $+0.248$] & Yes \\
\midrule
Single-direction cells: $\Delta\si$ & $+0.069$ [$+0.019$, $+0.119$] & Yes \\
Mismatch labels: $\Delta\si$ & $+0.213$ [$+0.163$, $+0.261$] & Yes \\
Modality dropout: audio-following $\Delta$ (points) & $-17.6$ [$-23.2$, $-11.9$] & Yes \\
Modality dropout: $\Delta\si$ (descriptive) & $+0.195$ [$+0.134$, $+0.255$] & --- \\
\bottomrule
\end{tabular}
\end{table}

\textbf{Primary results.}
The repair effect holds on the confirmation set, where all seven pre-registered tests meet their criteria (Table~\ref{tab:conf}).
These results confirm the direction and the statistical reliability of the repair.
Because the confirmation families played no role in selecting the recipe, the $59.9\%$ reduction in Section~\ref{sec:main_results} is not an artifact of tuning on the development set.

\begin{figure}[t]
\centering
\includegraphics{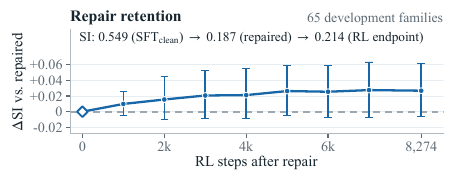}
\caption{Change in Audio-question $\si$ from the repaired model during subsequent RL post-training, with pointwise 95\% paired family-cluster bootstrap intervals.}
\label{fig:retention_steps}
\end{figure}

\textbf{Retention criteria.}
Subsequent RL leaves the index unchanged and further improves audio-following, which meets both retention criteria (Table~\ref{tab:conf}).
H4 requires the RL endpoint to retain at least $80\%$ of the repair gain $B$ in the index, which holds when the interval of $D - 0.2B$ lies below zero.
G2 applies the same $80\%$ criterion to the audio-following rates $p_S$, $p_F$, and $p_R$ over all conflict cells and holds when the lower confidence bound of $p_R - 0.8\,p_F - 0.2\,p_S$ exceeds zero.
The development index remains near its repaired value throughout the RL run of Section~\ref{sec:retention} (Figure~\ref{fig:retention_steps}), while IntentBench accuracy improves (Figure~\ref{fig:retention}).
The suppression therefore does not depend on stopping training at the repaired checkpoint.

\textbf{Equivalence band of H2.}
H2 confirms that the repair shrinks the mean image effect below a tenth of its starting magnitude.
The pre-registration text specifies a $\pm20\%$ equivalence band for H2, whereas the frozen analysis script implements the stricter $\pm10\%$ band that is reported in Table~\ref{tab:conf}.
The confidence intervals satisfy both bands.

\textbf{Construction controls.}
In all three pre-registered exploratory comparisons, each control underperforms the complete-pool LoRA reference, as specified before the evaluation (lower block of Table~\ref{tab:conf}).
For the modality-dropout control, the pre-registered comparison concerns the generated answers, and its change in the index is a descriptive result.
For the mismatch-label control, the pre-specified evaluation of the generated answers finds an explicit mismatch verdict in about a third of the conflict-cell outputs, and nearly all of these verdicts are among the outputs that cannot be parsed as \texttt{yes} or \texttt{no}.
Counting these outputs as failures places the control below the starting checkpoint in audio-following (Table~\ref{tab:gen_decomp}).
This is the behavior anticipated in Section~\ref{sec:relabel} for a label that does not depend on the designated modality.

\textbf{Unparseable outputs.}
The lead of the complete grid in free generation (Section~\ref{sec:dissection}) holds under all three treatments of unparseable outputs (Table~\ref{tab:gen_decomp}).
Most unparseable outputs of the other LoRA-trained models end in a malformed answer tag such as \texttt{\textless{}yes\textgreater{}Yes\textless{}/answer\textgreater{}}, which the pre-specified parser rejects, whereas most of those of the mismatch-label control are mismatch verdicts.
When these outputs count as failures, the single-direction control trails the reference both because more of its outputs are unparseable and because more of its answers follow the image.
When they are excluded, every control remains below the reference in audio-following, with intervals that exclude zero.
As a third treatment, we read the text after \texttt{\textless{}/think\textgreater{}} of an unparseable output and accept it as an answer if it contains \texttt{yes} but not \texttt{no}, or \texttt{no} but not \texttt{yes}.
This recovers most unparseable outputs of the reference and of the single-direction control, and every control again remains below the reference.
This analysis of the unparseable outputs was not pre-registered and is descriptive.

\begin{table}[!ht]
\caption{Generated answers on the 510 confirmation conflict cells, as family-averaged shares in percent of answers that follow the audio, answers that follow the image, and unparseable outputs.
The first numeric column is the pre-registered audio-following rate, in which unparseable outputs count as failures.
The parseable columns exclude unparseable outputs, pooled over cells, and the lenient column accepts an unparseable output whose text after \texttt{\textless{}/think\textgreater{}} contains only one of \texttt{yes} and \texttt{no}, and counts the other unparseable outputs as failures.
The complete grid is the LoRA reference of Figure~\ref{fig:controls}, and differences to it are paired, with 95\% family-cluster bootstrap intervals.}
\label{tab:gen_decomp}
\centering
\footnotesize
\setlength{\tabcolsep}{2.5pt}
\begin{tabular}{lcccccc}
\toprule
 & \multicolumn{3}{c}{All conflict cells} & \multicolumn{2}{c}{Parseable outputs only} & Lenient \\
\cmidrule(lr){2-4}\cmidrule(lr){5-6}\cmidrule(l){7-7}
Training cells & Audio & Image & Unparseable & Audio & $\Delta$ to grid & $\Delta$ to grid \\
\midrule
SFT$_{\mathrm{clean}}$ (no training) & 42.8 & 56.6 & 0.6 & 43.0 & $-24.0$ [$-30.7$, $-17.3$] & $-22.1$ [$-28.5$, $-15.6$] \\
Complete $2\times2$ grid & 61.3 & 30.5 & 8.2 & 67.0 & --- & --- \\
Single-direction cells & 45.1 & 36.9 & 18.0 & 55.1 & $-11.8$ [$-17.3$, $-6.3$] & $-12.3$ [$-17.8$, $-7.0$] \\
Mismatch labels & 33.6 & 29.9 & 36.5 & 52.8 & $-14.2$ [$-19.7$, $-8.6$] & $-31.2$ [$-36.1$, $-26.4$] \\
Modality dropout & 43.8 & 41.6 & 14.6 & 51.1 & $-15.8$ [$-22.0$, $-9.6$] & $-14.5$ [$-20.5$, $-8.6$] \\
\bottomrule
\end{tabular}
\end{table}

\section{Zero-Shot AVQA Protocol}
\label{app:avqa}

This appendix supports the zero-shot result of Section~\ref{sec:transfer} by showing that all three pre-registered criteria are met on AVQA.
For the AVQA diagnostic set, we apply the construction rules of Section~\ref{sec:design} to the items of \texttt{OmniInstruct\_v1} \citep{omnibench} whose source field is AVQA.
The questions of these items originate from AVQA \citep{avqa} and their clips from VGGSound \citep{vggsound}.
We add a global media-uniqueness constraint, because AVQA reuses source clips across questions and a single connected component of shared media would otherwise span a large share of the families.
Enforcing this constraint removes one family and leaves no media shared across families (Table~\ref{tab:index}).
None of these clips appears in any training configuration.
Three criteria and their thresholds were registered before the frozen checkpoints were evaluated, and Table~\ref{tab:avqa} reports them.

\begin{table}[!ht]
\caption{Pre-registered AVQA criteria, evaluated once on 104 families after the checkpoints were frozen. Intervals are 95\% family-cluster bootstrap CIs over 8{,}000 resamples, and C3 is paired against SFT$_{\mathrm{clean}}$.
The per-pair $\Delta\si$ row follows the registered definition, which averages the signed ratio $\ev/(|\ea|+|\ev|)$ over pairs.
The family-level $\si$ of Section~\ref{sec:design}, which Table~\ref{tab:avqa_full} reports, changes by $-0.174$ [$-0.210$, $-0.140$].
The generation rows use 40 of the 104 families (160 conflict cells) on AVQA and the 65 development Audio families on MUSIC-AVQA.}
\label{tab:avqa}
\centering
\small
\begin{tabular}{llcc}
\toprule
 & Statistic & Result & Met \\
\midrule
C1 answerable & SFT$_{\mathrm{clean}}$ matched-cell accuracy $>$ chance & $68.3\%$ [$64.2$, $72.4$] & Yes \\
C2 replicates & SFT$_{\mathrm{clean}}$ $\ev > 0$ & $+0.852$ [$+0.694$, $+1.008$] & Yes \\
C3 repair transfers & $\Delta\ev < 0$ & $-0.529$ [$-0.664$, $-0.399$] & Yes \\
 & per-pair $\Delta\si < 0$ & $-0.104$ [$-0.167$, $-0.040$] & Yes \\
 & matched-cell accuracy drop $\le 2$ points & $68.3\% \to 82.2\%$ & Yes \\
\midrule
descriptive & conflict audio-following (gen.) & $44.4\% \to 62.5\%$ & --- \\
 & same quantity on MUSIC-AVQA & $43.1\% \to 61.2\%$ & --- \\
\bottomrule
\end{tabular}
\end{table}

Before the repair, the index is lower on AVQA than on MUSIC-AVQA because the audio effect of SFT$_{\mathrm{clean}}$ is larger ($+1.56$ versus $+0.90$) and a similar image effect accounts for a smaller share (Table~\ref{tab:avqa_full}).
The image effect is nevertheless significantly positive before the repair, and both the image effect and the index decrease afterward.
The repair reduces the same reliance on the image for questions about everyday sounds, outside the music domain of its training cells.

\begin{table}[!ht]
\caption{Audio-question diagnostic on the 104 AVQA families under the native prompt. Intervals are 95\% family-cluster bootstrap CIs over 8{,}000 resamples.}
\label{tab:avqa_full}
\centering
\footnotesize
\setlength{\tabcolsep}{4pt}
\begin{tabular}{lccc}
\toprule
Model & $\ea$ [95\% CI] & $\ev$ [95\% CI] & $\si$ [95\% CI] \\
\midrule
Base & $+2.911$ [$+2.564$, $+3.286$] & $+0.950$ [$+0.764$, $+1.149$] & 0.292 [0.253, 0.330] \\
HumanOmniV2 & $+3.381$ [$+2.970$, $+3.785$] & $+1.001$ [$+0.801$, $+1.199$] & 0.302 [0.258, 0.347] \\
SFT$_{\mathrm{clean}}$ & $+1.556$ [$+1.337$, $+1.776$] & $+0.852$ [$+0.694$, $+1.008$] & 0.392 [0.349, 0.436] \\
Repaired (full) & $+1.522$ [$+1.347$, $+1.693$] & $\mathbf{+0.323}$ [$+0.247$, $+0.402$] & \textbf{0.218} [0.186, 0.251] \\
RL endpoint & $+2.044$ [$+1.806$, $+2.276$] & $+0.466$ [$+0.359$, $+0.576$] & 0.238 [0.201, 0.274] \\
\bottomrule
\end{tabular}
\end{table}

\newpage
\section{Zero-Shot AVHBench Protocol}
\label{app:avh}

This appendix details the AVHBench evaluation behind Sections~\ref{sec:main_results} and~\ref{sec:transfer} and traces the audio-task gain to audible objects.
We evaluate the three yes/no tasks of AVHBench \citep{avhbench}, which comprise 5{,}302 questions about AudioCaps and VALOR clips in their original and audio-swapped versions. These questions reference 2{,}092 distinct video identifiers in our evaluation files, while the benchmark reports 2{,}136 videos across its four tasks.
Video-driven audio hallucination asks whether a visible object is audible (2{,}290 questions).
Audio-driven video hallucination asks whether an audible object is visible (1{,}136 questions), and audio-visual matching contributes another 1{,}876 questions.
Each task has balanced labels.

We use the IntentBench evaluation pipeline for the Qwen checkpoints.
Video and audio are supplied as separate inputs, with the HumanOmniV2 system prompt and output format.
The predicted answer is the first yes/no token within the \texttt{<answer>} tags.
Unparseable outputs count as errors and make up at most $0.2\%$ of outputs for every Qwen checkpoint.
MiniCPM-o-2.6 is scored as in its diagnostic, with the neutral prompt and the first-token yes/no margin (Appendix~\ref{app:details}).
Its diagnostic interface takes a single image, so the model receives the middle frame of each video together with the full audio track.
Accuracy and the paired difference of Section~\ref{sec:main_results} are averaged over questions.
The interval of this difference uses a paired bootstrap over video clusters with 10{,}000 resamples, retaining the same per-question weighting, so the gain equals the difference between the unrounded accuracies.

\begin{table}[!ht]
\caption{Zero-shot AVHBench accuracy (\%) by task, with the yes-rate (\%) in parentheses. Tasks are video-driven audio hallucination (audio hall.), audio-driven video hallucination (video hall.), and audio-visual matching. The table covers all 5{,}302 questions, and ``Overall'' is the question-weighted mean over the three tasks. The two model families use different answer readouts and are not directly comparable to each other, so each is compared with its own starting checkpoint. Of the audio-hallucination answers that change from \texttt{no} to \texttt{yes} and from \texttt{yes} to \texttt{no} between the starting checkpoint and the RL endpoint, $69\%$ and $64\%$, respectively, are correct at the RL endpoint.}
\label{tab:avh}
\centering
\footnotesize
\setlength{\tabcolsep}{3pt}
\begin{tabular}{llcccc}
\toprule
Model & Readout & Audio hall. & Video hall. & AV matching & Overall \\
\midrule
Qwen2.5-Omni SFT$_{\mathrm{clean}}$ & gen. & 61.9 (37.2) & 81.5 (48.6) & 55.4 (62.3) & 63.8 \\
\quad + repair (full) & gen. & 67.7 (44.1) & 82.0 (47.9) & 62.6 (58.6) & 69.0 \\
\quad + subsequent RL & gen. & \textbf{73.2} (52.1) & \textbf{82.2} (54.2) & \textbf{62.7} (65.8) & \textbf{71.4} \\
\midrule
MiniCPM-o-2.6 (base) & margin & 78.7 (44.2) & 76.0 (31.8) & 64.0 (19.1) & 72.9 \\
\quad + repair (LoRA) & margin & \textbf{80.2} (41.1) & \textbf{78.0} (35.2) & \textbf{64.8} (16.6) & \textbf{74.3} \\
\bottomrule
\end{tabular}
\end{table}

The repair improves the two tasks that require listening, and subsequent RL extends this gain only on video-driven audio hallucination (Table~\ref{tab:avh}).
On audio-driven video hallucination and audio-visual matching, the RL endpoint remains within a few tenths of a point of the repaired model.
The gains of both stages concentrate on the listening tasks, where the diagnostic locates the shortcut.

The audio-task gain primarily reflects improved detection of previously missed audible objects.
Accuracy on audible objects rises at each stage (Section~\ref{sec:transfer}), whereas accuracy on silent objects falls by less than four points.
Within the audio task, answers that change from \texttt{no} at the starting checkpoint to \texttt{yes} at the RL endpoint outnumber the reverse changes by almost three to one.
Because most changed answers are corrections in both directions, the rising yes-rate in Table~\ref{tab:avh} mainly reflects the detection of audible objects rather than a general preference for \texttt{yes}.

Answer accuracy improves even though most outputs still omit an explicit audio description.
Empty audio descriptions (\texttt{<not\_available>}) occur in nearly all outputs of the starting checkpoint and still in more than four in five after the repair and after RL.
This is expected under answer-token supervision, which sets no target for the generated description (Section~\ref{sec:supervision}).

The LoRA repair also improves the accuracy of MiniCPM-o-2.6 over its base model, most on the two hallucination tasks.
In this model family, the yes-rate falls on the audio task and on audio-visual matching as accuracy rises (Table~\ref{tab:avh}).
The cross-model gain therefore does not stem from a general shift toward \texttt{yes}.

Table~\ref{tab:avh_compare} in Section~\ref{sec:transfer} compares the Qwen results with published methods on the same backbone.

\section{Generated Answers on Conflict Cells}
\label{app:qual}

This appendix adds eight conflict cells to the case study of Section~\ref{sec:case}.
All outputs use greedy decoding under the native prompt (Appendix~\ref{app:prompts}), and the text in each box is quoted from the model's output and truncated.
Questions keep the wording of the datasets.

\begin{figure}[!ht]
\centering
\includegraphics[width=\textwidth]{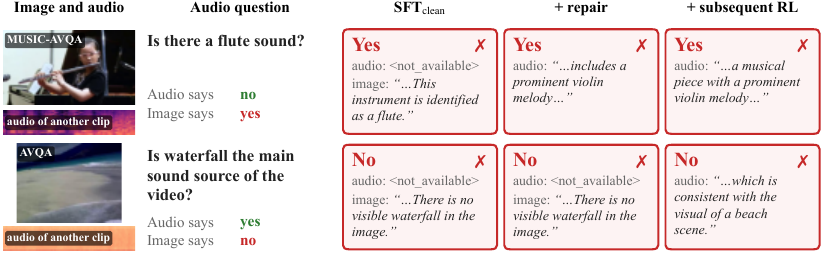}
\caption{Two conflict cells in the format of Figure~\ref{fig:qual}. All three checkpoints follow the image.}
\label{fig:qual_extra}
\end{figure}

Figure~\ref{fig:qual_extra} shows two cells in the format of Figure~\ref{fig:qual} on which all three checkpoints follow the image.
On the MUSIC-AVQA cell, the audio is an ensemble piece, and the repaired model and the RL endpoint describe its violin melody but still answer from the image.
On the AVQA cell, the RL endpoint describes the audio through the beach scene in the image.

\begin{figure}[t]
\centering
\includegraphics[width=\textwidth]{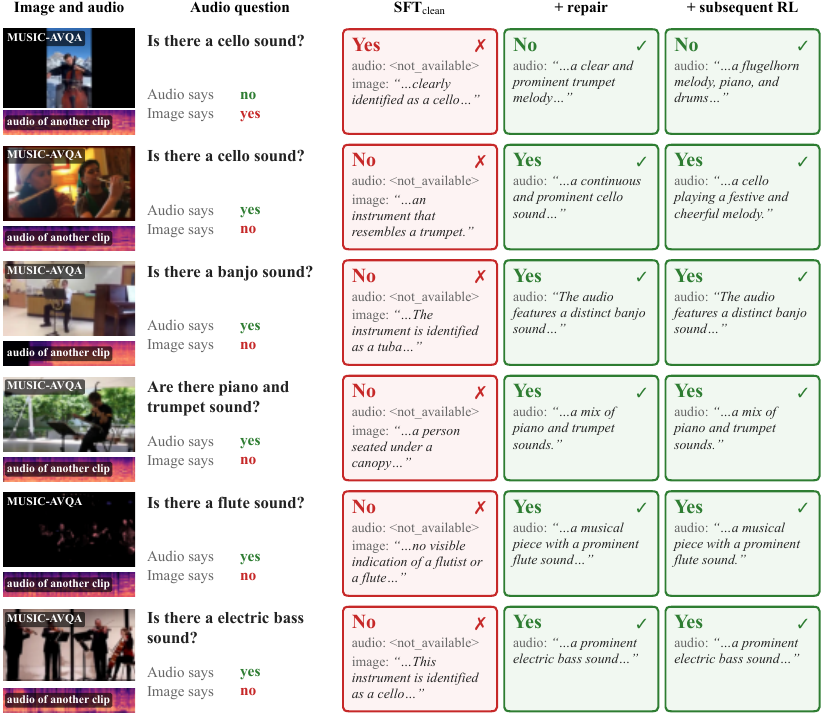}
\caption{Six conflict cells in the format of Figure~\ref{fig:qual}, four from development pairs and two, in rows three and four, from the confirmation set.
Rows one and two are the two directions of the same pair, and the remaining rows show cells from four other pairs.
Green boxes indicate outputs that follow the audio, and red boxes indicate outputs that follow the image.}
\label{fig:qual_more}
\end{figure}

Figure~\ref{fig:qual_more} adds six successful cells, four from development pairs and two, in rows three and four, from the confirmation set.
On 67 of the 260 development conflict cells, the starting checkpoint follows the image, whereas both the repaired model and the RL endpoint follow the audio.
The four development rows are drawn from these, and the two confirmation rows follow the same pattern.
In Figure~\ref{fig:qual_more}, the audio description of SFT$_{\mathrm{clean}}$ is empty on every cell and its answer follows its image description, whereas the repaired model and the RL endpoint state what the audio contains.

\section{Loss and Parameterization Ablations}
\label{app:ablation}

This appendix supports the attribution of the gain to the training cells in Section~\ref{sec:dissection}.
Table~\ref{tab:ablation} lists the variants that underlie the loss and parameterization paragraph of Section~\ref{sec:dissection}, together with the matched-cells-only control.
Every run is evaluated on the 65 development Audio families and follows the recipe of Appendix~\ref{app:details}, except the two runs marked $^{\dagger}$ in Table~\ref{tab:ablation}, which are trained with cross-entropy on an earlier and smaller pool.
$\Delta\si$ is the paired difference from the starting checkpoint under direct scoring, with a family-cluster bootstrap interval.
For audio-following, we compute within each family the proportion of all conflict cells whose generated answers follow the audio, and then average these proportions across families.

\begin{table}[!ht]
\caption{Loss and parameterization variants, development families.
The reference row is the LoRA recipe on the complete pool, which is also the reference of Figure~\ref{fig:controls}.
The seed row reports the range over four seeds of that reference, each with an interval that excludes zero, and full fine-tuning differs from the reference by $-0.007$ [$-0.053$, $+0.042$] in paired $\Delta\si$.
The matched-cells-only control differs from SFT$_{\mathrm{clean}}$ by $-4.6$ [$-12.3$, $+3.5$] points in audio-following.
$^{\dagger}$Trained with cross-entropy on an earlier and smaller pool.}
\label{tab:ablation}
\centering
\footnotesize
\setlength{\tabcolsep}{4.5pt}
\begin{tabular}{llcc}
\toprule
Variant & Training cells & $\Delta\si$ [95\% CI] & Audio-Following, \% (gen.) \\
\midrule
SFT$_{\mathrm{clean}}$ (no training) & --- & --- & 43.1 \\
Hinge, LoRA (reference) & complete $2\times2$ & $-0.356$ [$-0.427$, $-0.285$] & 60.8 \\
Hinge, full fine-tuning & complete $2\times2$ & $-0.362$ [$-0.430$, $-0.293$] & 61.2 \\
Cross-entropy, LoRA & complete $2\times2$ & $-0.317$ [$-0.377$, $-0.258$] & 61.9 \\
Hinge, LoRA, four seeds & complete $2\times2$ & $-0.356$ to $-0.296$ & --- \\
Cross-entropy, LoRA$^{\dagger}$ & complete $2\times2$ & $-0.297$ [$-0.364$, $-0.232$] & --- \\
Matched cells only, LoRA$^{\dagger}$ & two original cells & $-0.101$ [$-0.166$, $-0.038$] & 38.5 [32.3, 44.6] \\
\bottomrule
\end{tabular}
\end{table}

The loss and the parameterization have little effect on the index reduction.
Cross-entropy achieves $89\%$ of the index reduction obtained with the hinge loss under the same recipe, and full fine-tuning and LoRA reach almost the same $\Delta\si$.

Removing the counterfactual cells has a much larger effect.
The matched-cells-only control achieves less than a third of the index reduction of the reference.
Relative to cross-entropy training on all cells of the same earlier pool, it achieves about a third of the reduction, as stated in Section~\ref{sec:dissection}.
The control's audio-following does not differ detectably from that of the starting checkpoint (Table~\ref{tab:ablation}).
Without counterfactual cells, the image predicts the label as well as the audio does (Section~\ref{sec:relabel}).

\section{Cross-Model Generalization on MiniCPM-o-2.6}
\label{app:xmodel}

Table~\ref{tab:xmodel} gives the Audio-question diagnostic behind the cross-model result of Section~\ref{sec:main_results}.
The LoRA repair of MiniCPM-o-2.6 uses the same training cells as the Qwen recipe (Appendix~\ref{app:details}) and reduces the index by about one half on both the development and the confirmation set.
The image effect falls to about one tenth of its value before the repair on both sets, whereas the audio effect decreases by about one fifth.
The lower index reflects the removal of most of the image effect, while most of the audio effect remains.

\begin{table}[H]
\caption{Audio-question diagnostic of MiniCPM-o-2.6 before and after the repair, under the neutral prompt and the first-token margin.
Arrows go from the base model to the repaired model, and $\Delta\si$ is the paired change with its 95\% family-cluster bootstrap interval over 10{,}000 resamples.}
\label{tab:xmodel}
\centering
\footnotesize
\setlength{\tabcolsep}{4pt}
\begin{tabular}{lcccc}
\toprule
Set (Audio families) & $\ea$ & $\ev$ & $\si$ & $\Delta\si$ [95\% CI] \\
\midrule
Development (65) & $+1.57 \to +1.27$ & $+1.88 \to +0.20$ & $0.554 \to 0.261$ & $-0.293$ [$-0.361$, $-0.223$] \\
Confirmation (128) & $+1.17 \to +0.92$ & $+1.68 \to +0.17$ & $0.604 \to 0.297$ & $-0.307$ [$-0.365$, $-0.248$] \\
\bottomrule
\end{tabular}
\end{table}

\section{Prompts}
\label{app:prompts}

This appendix lists the prompts of all experiments.
Braces mark fields that are filled per sample, and the original spelling of each prompt is preserved.

\textbf{System prompt.}
All Qwen checkpoints under the native prompt, the subsequent GRPO runs, and the Qwen benchmark evaluations use the system prompt of HumanOmniV2 \citep{humanomniv2}.
\begin{quote}\small\ttfamily\raggedright
You are a helpful assistant. Your primary goal is to deeply analyze and interpret information from available various modalities (image, video, audio, text context) to answer questions with human-like depth and a clear, traceable thought process.\par\smallskip
Begin by thoroughly understanding the image, video, audio or other available context information, and then proceed with an in-depth analysis related to the question.\par\smallskip
In reasoning, It is encouraged to incorporate self-reflection and verification into your reasoning process. You are encouraged to review the image, video, audio, or other context information to ensure the answer accuracy.\par\smallskip
Provide your understanding of the image, video, and audio between the \textless{}context\textgreater{} \textless{}/context\textgreater{} tags, detail the reasoning between the \textless{}think\textgreater{} \textless{}/think\textgreater{} tags, and then give your final answer between the \textless{}answer\textgreater{} \textless{}/answer\textgreater{} tags.
\end{quote}

\textbf{Diagnostic and generation prompt.}
The diagnostic of Section~\ref{sec:design} and free generation give the model the image, the audio, and the following text.
\begin{quote}\small\ttfamily\raggedright
Here is the image, with the coresponding audio.\par
\{question\}\par
Possible answers: yes, no
\end{quote}
Direct scoring appends \texttt{\textless{}answer\textgreater{}} as the start of the assistant turn and reads $\log p(\texttt{yes}) - \log p(\texttt{no})$ at the next token.
For the own-reasoning index of Table~\ref{tab:main}, the model first generates up to 512 tokens greedily on the matched cell $(i_o, a_o)$.
We keep the text before its first \texttt{\textless{}answer\textgreater{}}, or the full text if there is none, and append \texttt{\textless{}answer\textgreater{}} to obtain the prefix shared by all four cells.
Free generation decodes greedily with up to 512 new tokens and takes the first \texttt{yes} or \texttt{no} after \texttt{\textless{}answer\textgreater{}}.
If there is no such match, the parser reads the answer region, or the last two lines of the output when it has no answer region.
It accepts this text if it contains affirmative words but no negative words, or negative words but no affirmative words.
Any remaining output counts as unparseable.
In the development evaluation, $1\%$ of the generated outputs of SFT$_{\mathrm{clean}}$, $8\%$ of those of the LoRA model, and $2\%$ of those of the fully fine-tuned model are unparseable and count as failures.
The neutral prompt omits the system prompt and the answer prefix, and scoring uses the first assistant token.
In both the diagnostic and AVHBench, MiniCPM-o receives the same text after its media markers \texttt{(\textless{}image\textgreater{}./\textless{}/image\textgreater{})(\textless{}audio\textgreater{}./\textless{}/audio\textgreater{})}.

\textbf{Training prompt.}
DMC-Repair uses the same system prompt and text, except that the answer list contains all 42 answers of the training pool.
\begin{quote}\small\ttfamily\raggedright
Possible answers: accordion, acoustic\_guitar, bagpipe, banjo, bassoon, cello, clarinet, congas, drum, eight, electric\_bass, erhu, five, flute, four, guzheng, indoor, left, middle, more than ten, nine, no, one, outdoor, piano, pipa, right, saxophone, seven, simultaneously, six, suona, ten, three, trumpet, tuba, two, ukulele, violin, xylophone, yes, zero
\end{quote}
The score of a candidate is its mean token log-probability after \texttt{\textless{}answer\textgreater{}}, and the hinge loss of Equation~\ref{eq:hinge} compares the two candidates of each cell.
MiniCPM-o uses its own chat template with the media markers above and without a system prompt or answer prefix.

\textbf{Benchmark prompts.}
For the Qwen checkpoints, AVHBench and IntentBench are evaluated with the code of \citet{humanomniv2} and the system prompt above.
The user turn contains the video, its audio track, and the following text.
\begin{quote}\small\ttfamily\raggedright
Here is a \{data\_type\}, with the audio from the video.\par
\{question\}\{type\_template\}
\end{quote}
AVHBench uses \texttt{video} as the data type and the yes/no template below.
\begin{quote}\small\ttfamily\raggedright
Please answer Yes or No within the \textless{}answer\textgreater{} \textless{}/answer\textgreater{} tags.
\end{quote}
Multiple-choice questions list their options after the word \texttt{Options:} and use this template.
\begin{quote}\small\ttfamily\raggedright
Please provide only the single option letter (e.g., A, B, C, D, etc.) within the \textless{}answer\textgreater{} \textless{}/answer\textgreater{} tags.
\end{quote}
The other question types of IntentBench use the corresponding templates of the same code.
The GRPO runs use the same system prompt and templates, with a line break between the question and the template.

\section{Extended Related Work}
\label{app:related}

\textbf{Modality-aware RL objectives.}
SFFL \citep{sffl} pairs modality-specific reasoning traces with a modality-preference reward, OmniVideo-R1 \citep{omnivideo_r1} modifies query grounding and modality-attentive fusion, and MAPO \citep{mapo} reweights the policy gradient on modality-critical tokens alongside an attention-based auxiliary loss.
Caption-based self-verification \citep{visionaryr1} and modality-separated self-reward \citep{visionsr1} pursue the same goal for images.
DMC-Repair differs in that it changes which input predicts the training label.

\textbf{Diagnostics for modality dependence.}
\citet{mccd} introduce MUSIC-AVQA-R to test robustness to question rephrasing and propose logit-space debiasing, while Sensory PID \citep{sensorypid} decomposes modality contributions through partial information decomposition.
A probing study finds that audio information present in intermediate representations can be suppressed in later fusion layers \citep{avllm_see_hear}.
The Factorized Modality Diagnostic complements these analyses with a measurement of each modality's effect on individual answers.

\textbf{Hallucination mitigation and counterfactual training.}
Among preference methods, ACPO \citep{acpo} penalizes captions that ignore a swapped audio track, MoD-DPO \citep{moddpo} encourages invariance to irrelevant-modality changes, and OmniDPO \citep{omnidpo} combines text and multimodal preferences.
Training-free methods instead contrast or reweight modality branches at decoding time \citep{avcd,fmd,mad}.
Earlier counterfactual VQA work synthesizes examples by masking critical evidence and changing labels \citep{css}.
Table~\ref{tab:avh_compare} compares DMC-Repair with the published AVHBench results of several of these methods.

\end{document}